\documentclass[runningheads]{llncs}
\usepackage[T1]{fontenc}
\usepackage{graphicx}
\usepackage{caption}
\usepackage{subcaption}
\usepackage{hyperref}
\usepackage{amsfonts}
\usepackage[misc,geometry]{ifsym}
\usepackage{xcolor}
\begin{document}
\title{Towards Scenario-based Safety Validation for Autonomous Trains with Deep Generative Models}
\titlerunning{Towards Scenario-based Safety Validation with Deep Generative Models}
%
\author{Thomas Decker\inst{1,2} \Letter\and
Ananta R. Bhattarai\inst{1,3} \and
Michael Lebacher\inst{1}}
\authorrunning{T. Decker et al.}
%
\institute{Siemens AG, Munich, Germany \and
	Ludwig Maximilians Universität, Munich, Germany \and
	Technical University of Munich, Munich, Germany\\
	\email{\{thomas.decker;michael.lebacher\}@siemens.com; ananta.bhattarai@tum.de}}
\maketitle              
\begin{abstract}
Modern AI techniques open up ever-increasing possibilities for autonomous vehicles, but how to appropriately verify the reliability of such systems remains unclear. A common approach is to conduct safety validation based on a predefined Operational Design Domain (ODD) describing specific conditions under which a system under test is required to operate properly. However, collecting sufficient realistic test cases to ensure comprehensive ODD coverage is challenging. In this paper, we report our practical experiences regarding the utility of data simulation with deep generative models for scenario-based ODD validation. We consider the specific use case of a camera-based rail-scene segmentation system designed to support autonomous train operation. We demonstrate the capabilities of semantically editing railway scenes with deep generative models to make a limited amount of test data more representative. We also show how our approach helps to analyze the degree to which a system complies with typical ODD requirements. Specifically, we focus on evaluating proper operation under different lighting and weather conditions as well as while transitioning between them. 

\end{abstract}
\section{Introduction}
Artificial Intelligence (AI) enables technologies that can process vast amounts of data from various sources in real time and its potential for autonomous vehicles is progressively transforming the transportation industry. This is especially true for the railway domain, where driverless trains are associated with various economic and societal benefits \cite{trentesaux2018autonomous}. Moreover, fully automated trains are already in service for many years in constrained and well-controlled environments such as metro lines with platform screen doors \cite{flammini2022vision}. However, enabling operation in general open settings is significantly more demanding as trains are constantly required to perceive and interact with the current environment. While AI has shown promising capabilities in this regard \cite{tang2022literature}, it is still unclear how to rigorously assure the safety of such systems from a regulatory and legal perspective \cite{flammini2022vision}. A popular approach to conduct safety validation of automated vehicles is scenario-based testing \cite{riedmaier2020survey}. Ideally, fully automated trains are expected to handle any environmental conditions and even unexpected events in a safe and robust manner, but the resulting space of possible scenarios is infeasible to test globally. As a consequence, scenario-based testing is typically performed considering a predefined Operational Design Domain (ODD) \cite{koopman2018toward} which refers to all specific conditions under which a system is strictly required to behave properly including physical, geographical and regulatory constraints \cite{koopman2019many}. While there already exist proposals regarding ODD specifications for railway applications \cite{tonk2021towards}, collecting sufficient test cases covering all relevant aspects and systematically conducting appropriate evaluations still remains challenging. However, AI-powered data generation in the form of deep generative models has demonstrated remarkable capacities to realistically simulate complex data structures \cite{bond2021deep}. In this work, we propose a framework to systematically leverage deep generative models for scenario-based testing and summarize our practical experiences. Specifically, we create high-resolution image data with conditional Generative Adversarial Networks (cGANs) \cite{wang2018high} allowing us to fix high-level image contents, such as the position of rails or other objects while altering different ODD-related characteristics during simulation. In this way, we can make a limited number of test cases more representative for the purpose of safety validation. We further apply our approach to test a camera-based rail-scene segmentation model that is implemented via a deep neural network \cite{zhao2017pyramid}. Such systems enable accurate perception of the frontal environment which is crucial for safe train operation and obstacle detection \cite{ristic2021review}. We demonstrate how to perform a systematic model evaluation under natural perturbations like different lighting and weather conditions as well as while transitioning between them. Such an analysis complements classical robustness certification \cite{paterson2021deepcert,li2023sok} and provides an additional tool to validate system safety in a comprehensive way.

\section{Background}\label{background}
\subsubsection{GANs}
Generative adversarial networks (GANs) are a popular category of deep generative models that have been extensively studied in computer vision and demonstrate remarkable capabilities to simulate realistic images and videos \cite{liu2021generative}. GANs consist of two neural networks, a generator and a discriminator, that are trained in concert to create new samples resembling the training data. Conditional GANs (cGANs) are extended versions that allow controlling the properties of generated data via additional input arguments. For images, cGANs enable semantic editing, style translation or creating images with specific details \cite{pang2021image}.  

\subsubsection{Semantic Segmentation and RailSem19}
Semantic segmentation describes the task of dividing an image into semantically distinct sub-regions and assigning them a corresponding label. Deep neural networks attain state-of-the-art performances for this purpose and have also been applied in corresponding railway applications \cite{ristic2021review}. Such models are typically trained via labeled training data comprising images and matching ground truth semantic label masks. A popular metric to evaluate segmentation performance is the Intersection over Union (IoU) score ranging from 0 to 1, where a score of 1 indicates a perfect match between ground truth and the predicted regions and 0 means no overlap. RailSem19 \cite{zendel2019railsem19} is a publicly available dataset for semantic segmentation of railway scenes. It contains 8500 high-resolution images of real train and tram front views together with pixel-wise semantic labels corresponding to 19 different classes. The provided labels allow to distinguish a variety of different safety-critical objects such as rails, cars, humans or other on-rail vehicles. The dataset also covers various different operation environments, illuminations and weather conditions which all resemble typical components of ODD descriptions for railway applications \cite{tonk2021towards}.

\section{Proposed Methodology}
\label{pm}
\begin{figure}[t]
	\centering
	\includegraphics[width=1.\textwidth]{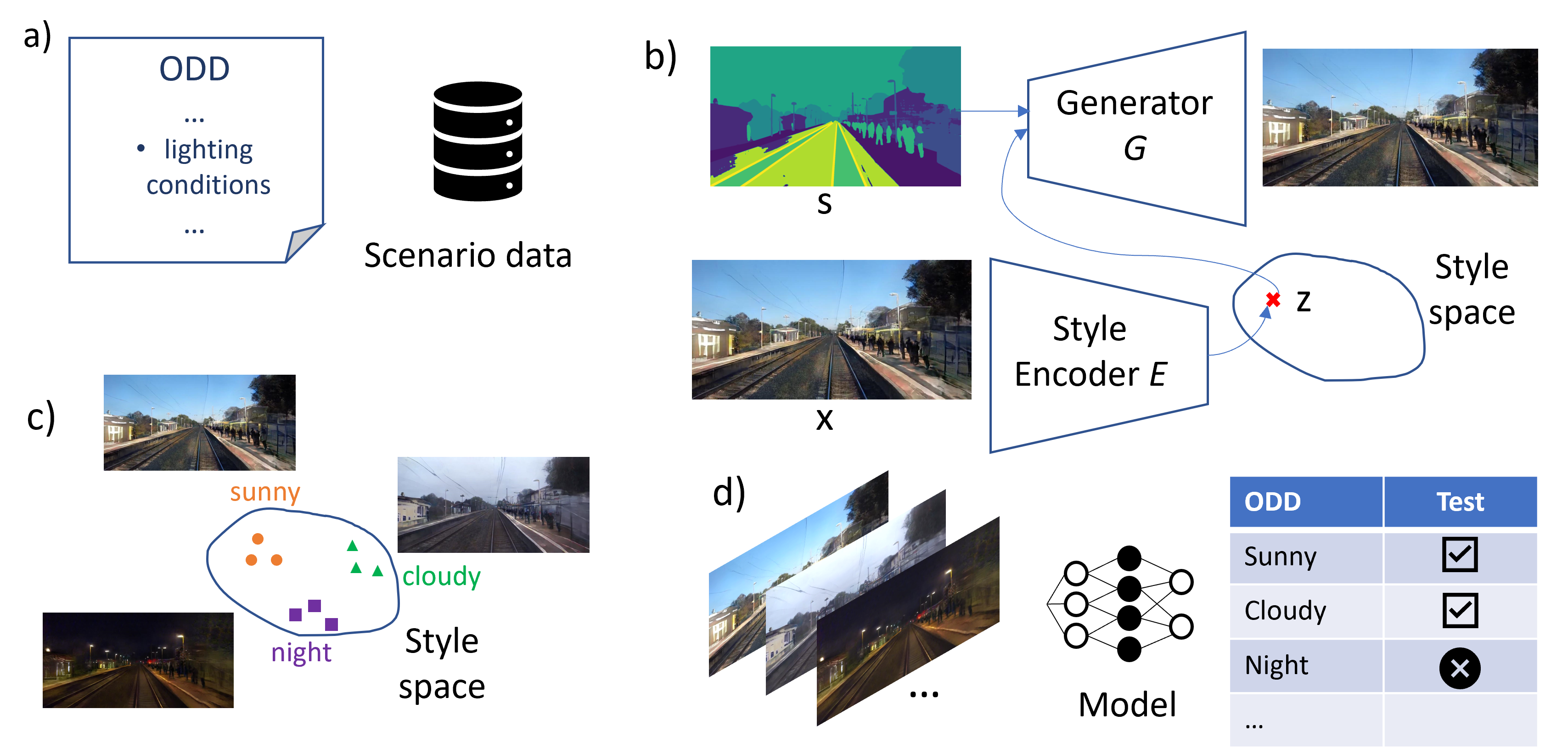}
	\caption{Proposed approach for scenario-based ODD validation with cGANs}\label{overview}
\end{figure}
The goal of our approach is to leverage deep generative models in a systematic way to validate if an AI-powered model fully complies with specific ODD requirements given only a limited amount of test cases. Our proposed methodology is illustrated in Fig \ref{overview}. As usual for safety validation, we suppose access to a predefined ODD description as well as a set of representative scenario data (a). In our use case, an ODD might among other things also require models to work well under changing lighting conditions and the extensive RailSem19 dataset provides corresponding scenarios. As a second step, we utilize the scenario data for training a cGAN to enable conditional generation of new relevant scenarios (b). In particular, we choose the pix2pixHD architecture \cite{wang2018high} that enables the creation of high-resolution images via a generator $G$ receiving two distinct inputs. First, the semantic structure of the desired image can be controlled by providing a semantic label mask $s$ informing $G$ where in the image specific objects or structures should appear. Second, a separate encoder network $E$ was designed to grasp the stylistic characteristics of different semantic categories. More precisely, $E$ encodes low-level details of regions in $x$ into low-dimensional feature vectors $z$ forming a numerical style space. This setup allows us to semantically manipulate a given scenario to increase test capacities and improve ODD coverage. To do so, we first run the trained encoder on all instances in the training set and save the resulting feature vectors. Following \cite{wang2018high}, we perform clustering on these features for each semantic category to localize ODD-related concepts in the style space (c). For instance, the cluster centers for the category Sky can encode styles such as sunshine, cloudiness or night. This enables us to synthesize new realistic images with identical high-level structures determined by $s$ but exhibiting different stylistic properties, like the same railway scene under varying lighting conditions. Moreover, we can also simulate continuous transitioning between two styles by interpolating the corresponding style encodings during image generation. To systematically test how well a model complies with an ODD requirement we can semantically manipulate available scenarios to exhibit specific properties and evaluate its effect on the model's performance (d). In the case of rail-scene segmentation this methodology allows us for instance to explicitly validate if a model works sufficiently well under sunny, cloudy or nighttime illumination.
\begin{figure}[!t]
\centering
    \begin{subfigure}[b]{0.32\textwidth}
        \centering
\includegraphics[width=\textwidth]{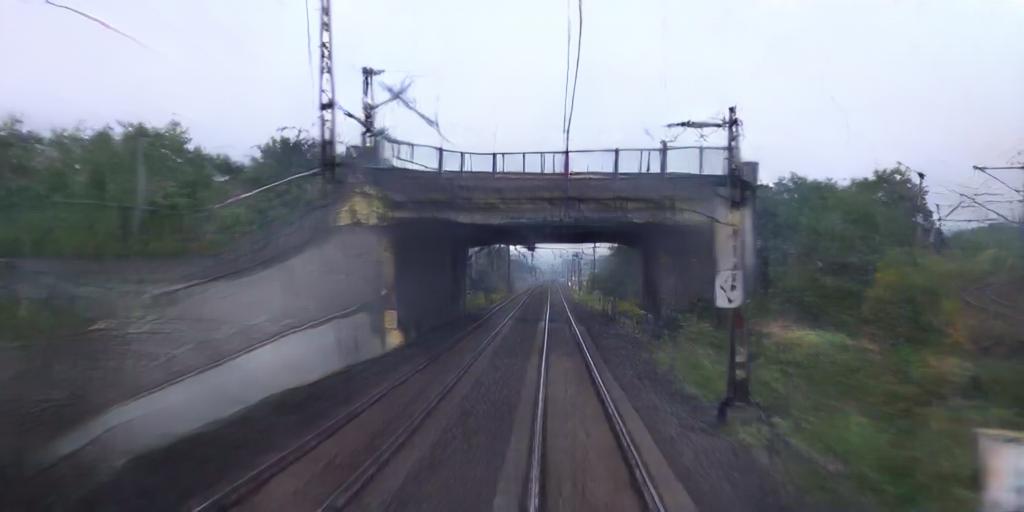}

    \end{subfigure}
    \begin{subfigure}[b]{0.32\textwidth}
        \centering
\includegraphics[width=\textwidth]{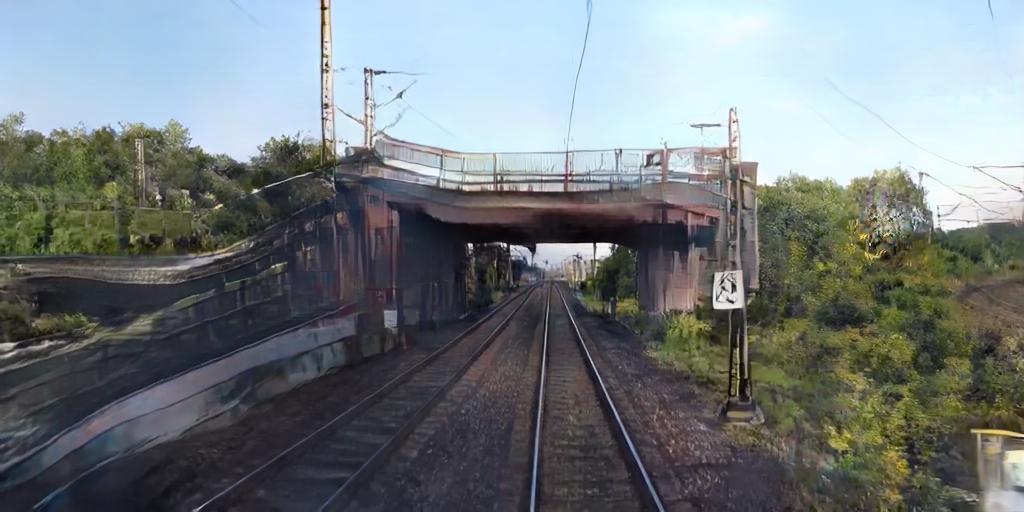}

    \end{subfigure}
    \begin{subfigure}[b]{0.32\textwidth}
        \centering
\includegraphics[width=\textwidth]{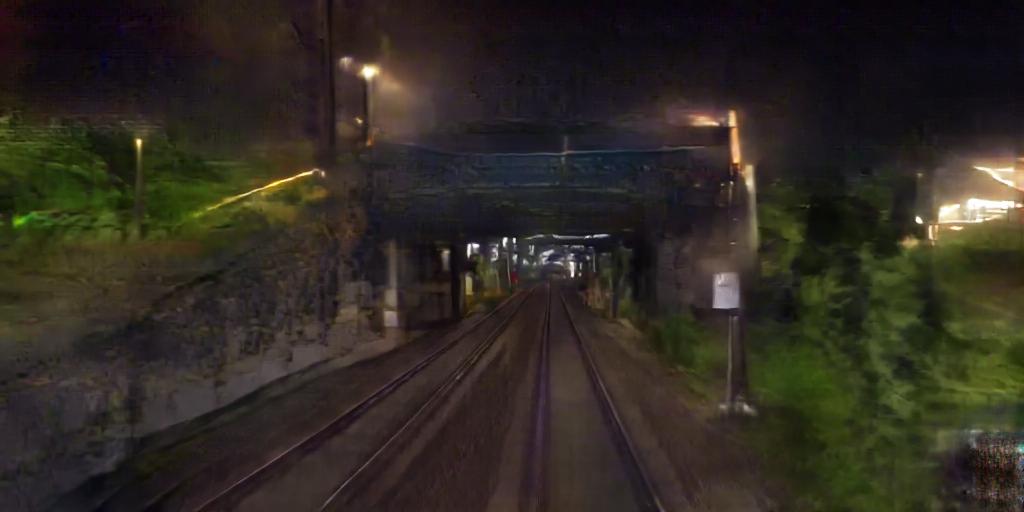}
    \end{subfigure}
    \caption{Styles represented by cluster centers of class Sky: cloudy, sunny and night.}
    \label{fig-1}
\end{figure}
\begin{figure}[!t]
\centering
    \begin{subfigure}[b]{0.3\textwidth}
        \centering
\includegraphics[width=\textwidth]{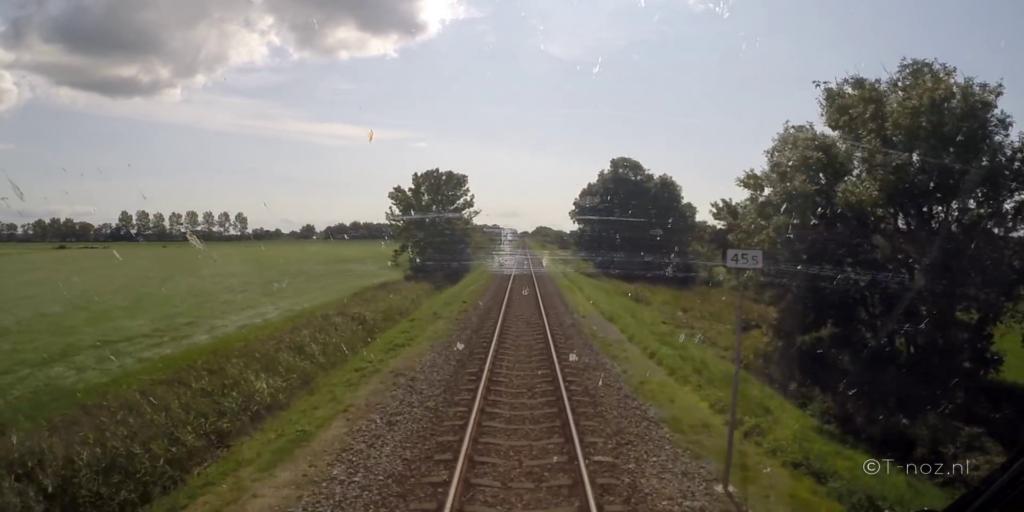}
        \caption*{Original image}
    \end{subfigure}
    \hspace{1cm}
        \begin{subfigure}[b]{0.3\textwidth}
        \centering
\includegraphics[width=\textwidth]{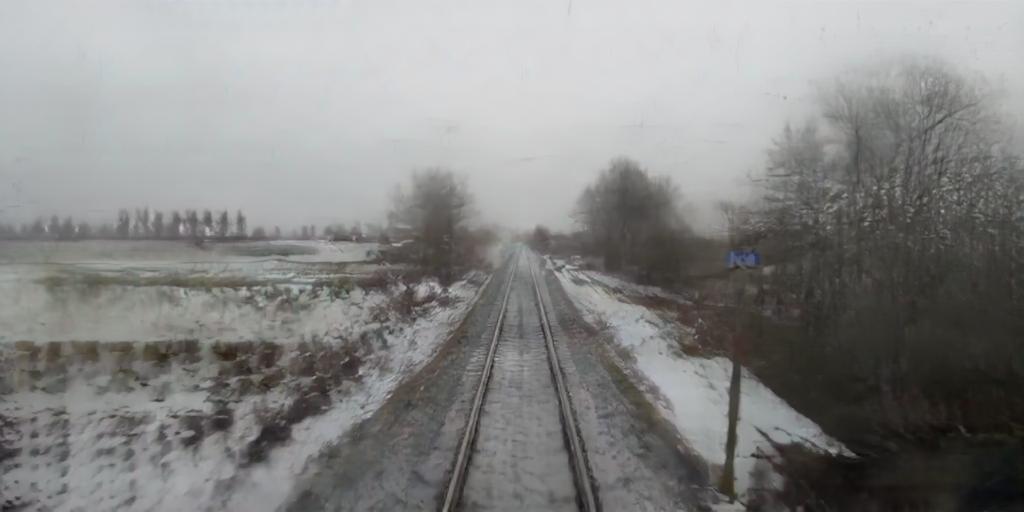}
        \caption*{Synthesized snow version}
    \end{subfigure}
    
    \begin{subfigure}[b]{0.19\textwidth}
        \centering
        \includegraphics[width=\textwidth]{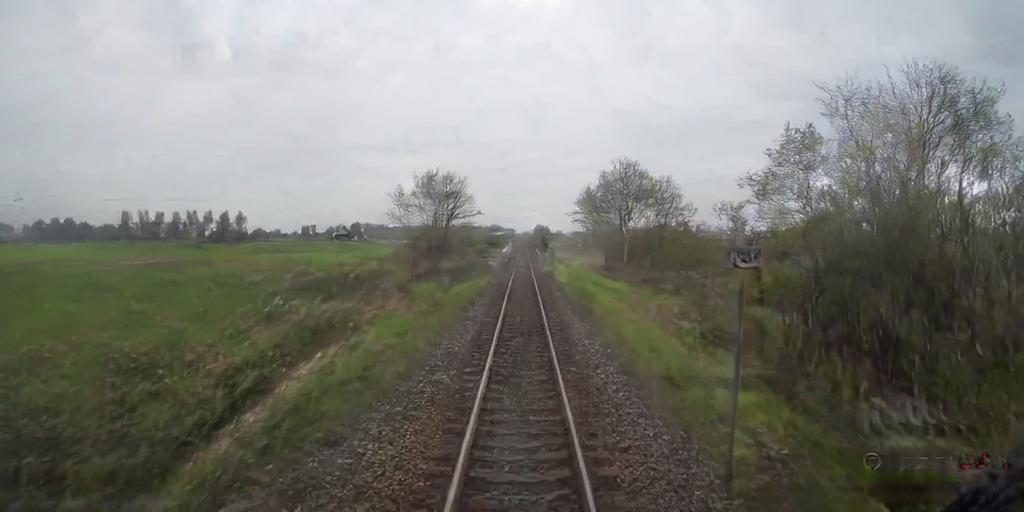}
        \caption*{Vegetation}
        \label{snow:a}
    \end{subfigure} 
    \begin{subfigure}[b]{0.19\textwidth}
        \centering
\includegraphics[width=\textwidth]{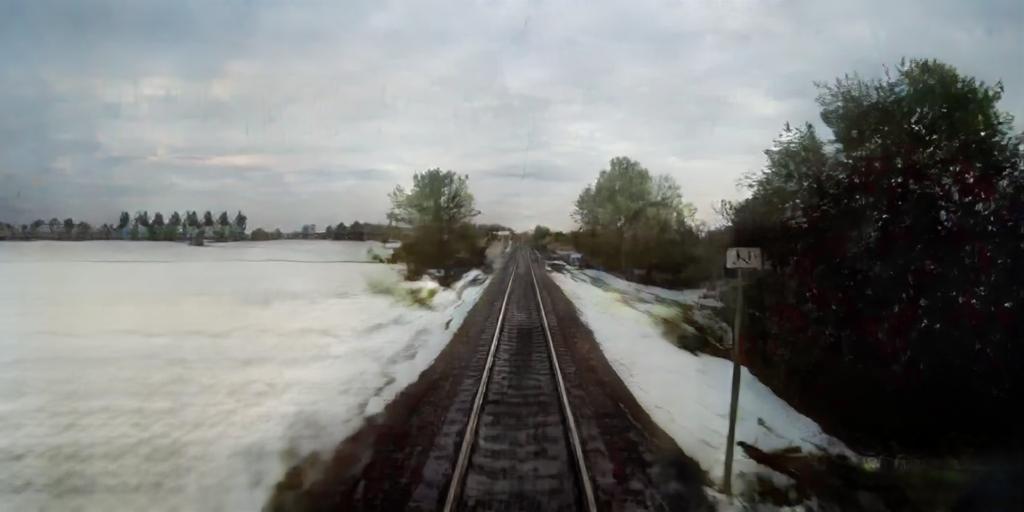}
        \caption*{Terrain}
        \label{snow:b}
    \end{subfigure}
    \begin{subfigure}[b]{0.19\textwidth}
        \centering
\includegraphics[width=\textwidth]{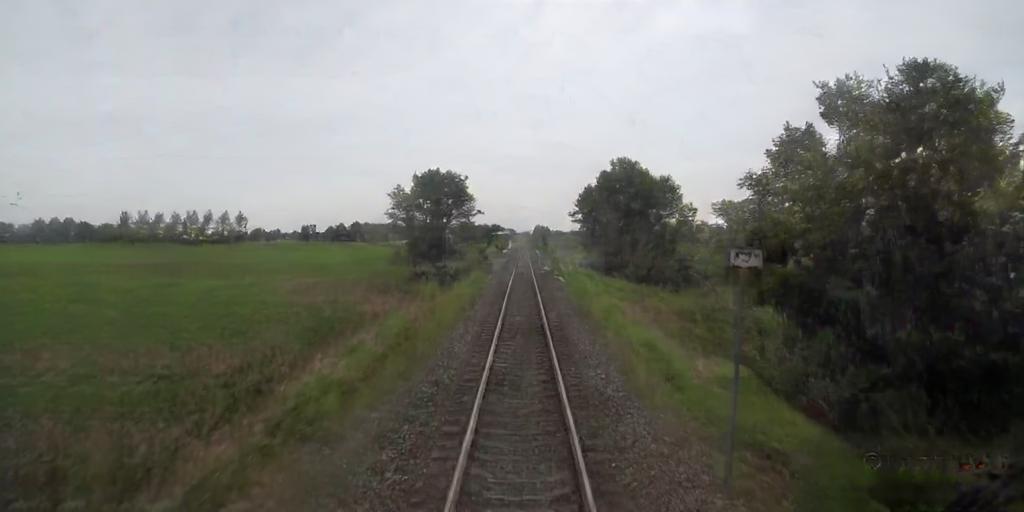}
        \caption*{Sky}
        \label{snow:c}
    \end{subfigure}
    \begin{subfigure}[b]{0.19\textwidth}
        \centering
\includegraphics[width=\textwidth]{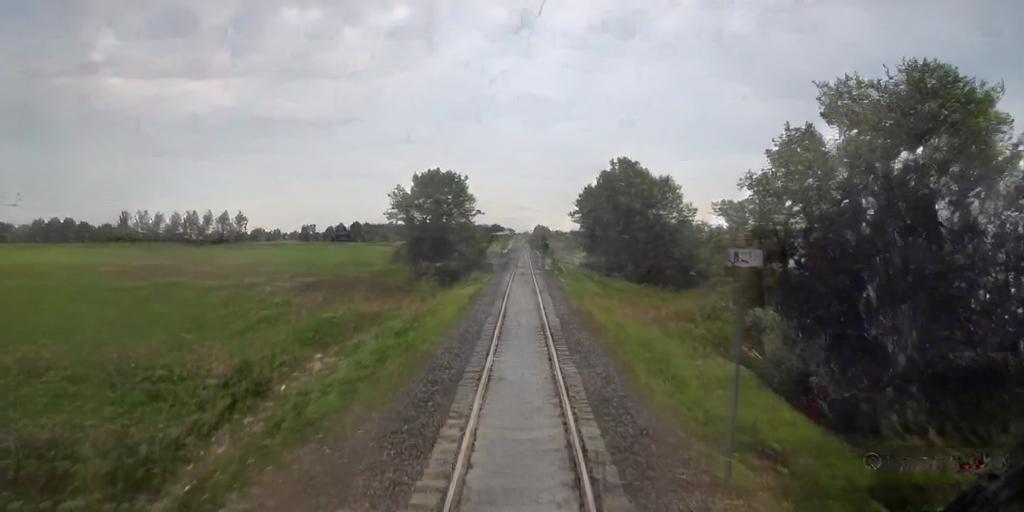}
        \caption*{Rail-track }
        \label{snow:d}
    \end{subfigure}
    \begin{subfigure}[b]{0.19\textwidth}
        \centering        \includegraphics[width=\textwidth]{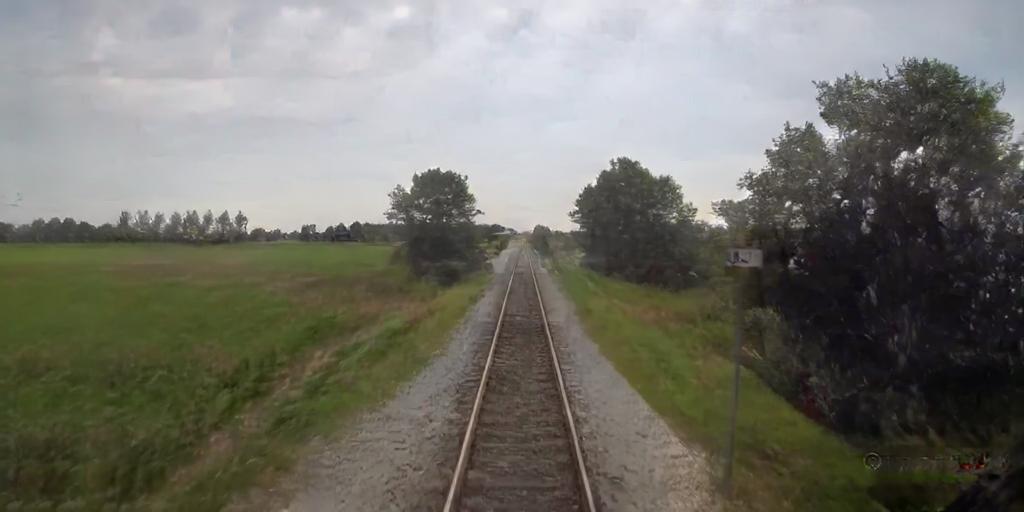}
        \caption*{Trackbed}
        \label{snow:e}
    \end{subfigure}  
    \caption{Synthesizing snowfall by altering features of different semantic categories.}   
    \label{snow-encoding}
\end{figure}
\begin{figure}[!t]
    \begin{minipage}{0.49\textwidth}
        \begin{subfigure}[b]{0.99\textwidth}
            \centering
\includegraphics[width=\textwidth]{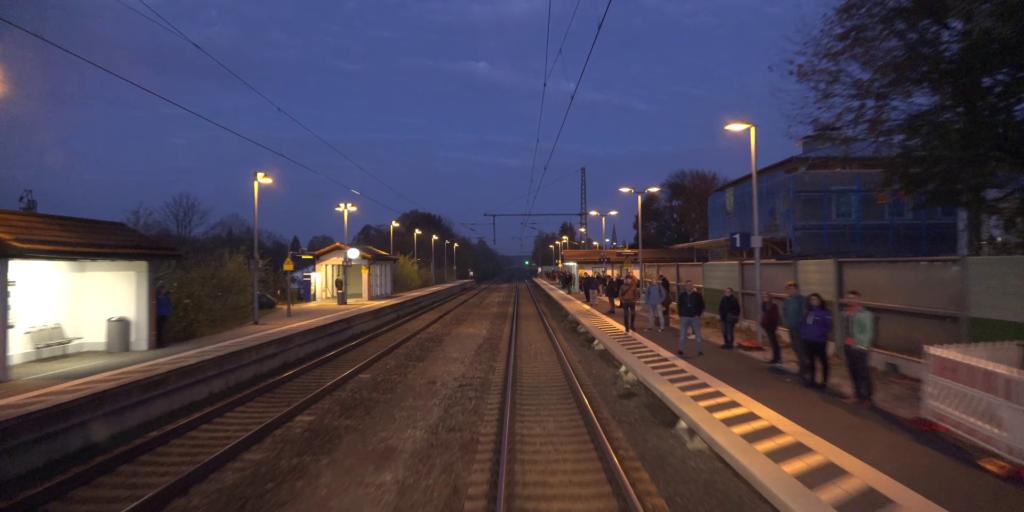}
            \caption*{Original Image}
            
        \end{subfigure}
    \end{minipage}
    \hfill
    \begin{minipage}{0.49\textwidth}
        \begin{subfigure}[b]{0.49\textwidth}
            \centering
\includegraphics[width=\textwidth]{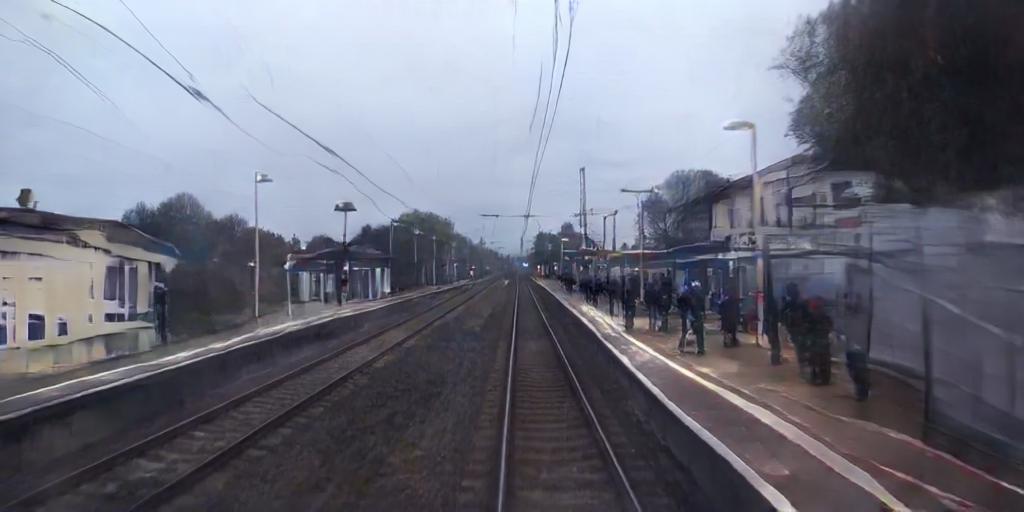}
            \caption*{Cloudy}
           
        \end{subfigure}
        \hfill
        \begin{subfigure}[b]{0.49\textwidth}
            \centering
\includegraphics[width=\textwidth]{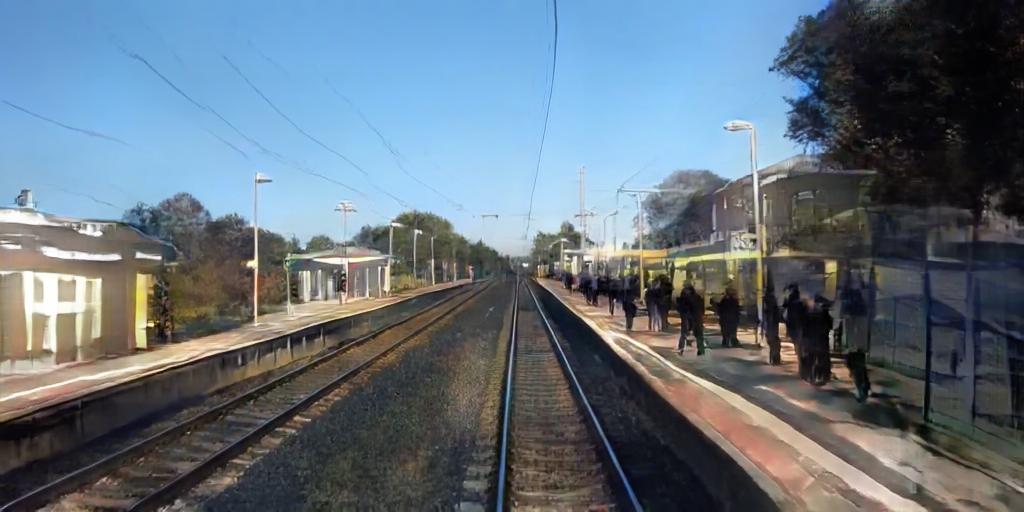}
            \caption*{Sunshine}
          
        \end{subfigure}
        \hfill
        \begin{subfigure}[b]{0.49\textwidth}
            \centering
\includegraphics[width=\textwidth]{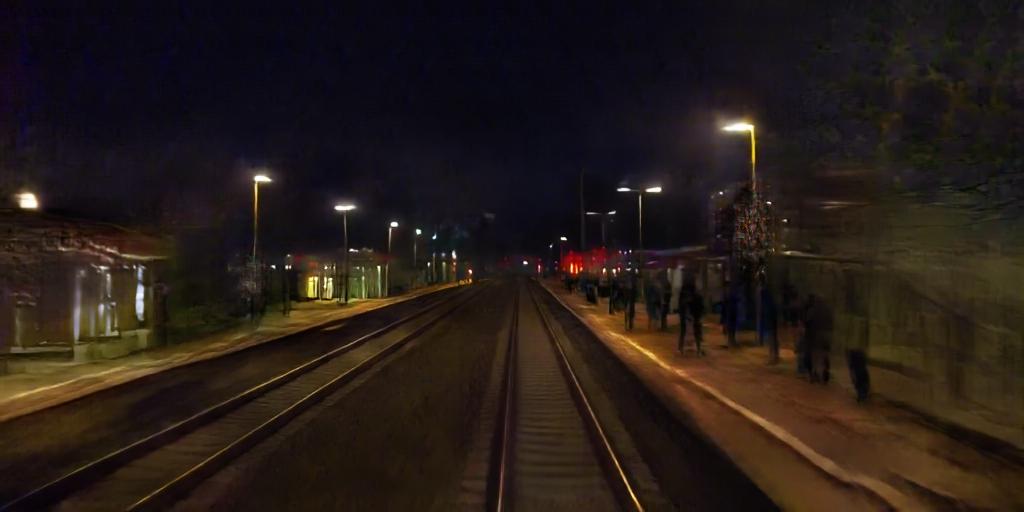}
            \caption*{Night}
            
        \end{subfigure}
        \hfill
        \begin{subfigure}[b]{0.49\textwidth}
            \centering
\includegraphics[width=\textwidth]{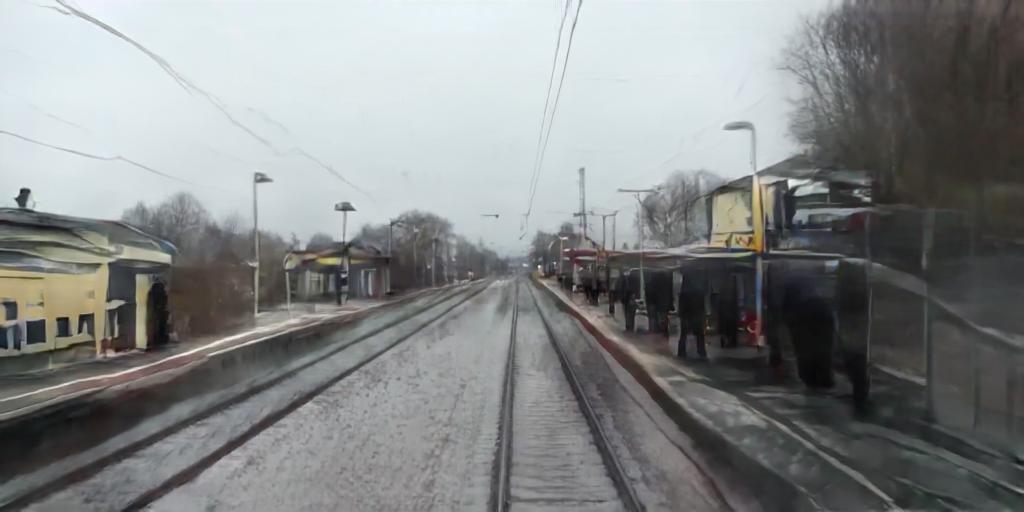}
            \caption*{Snow}
            
        \end{subfigure}
    \end{minipage}

    \begin{minipage}{0.49\textwidth}
        \begin{subfigure}[b]{0.99\textwidth}
            \centering
\includegraphics[width=\textwidth]{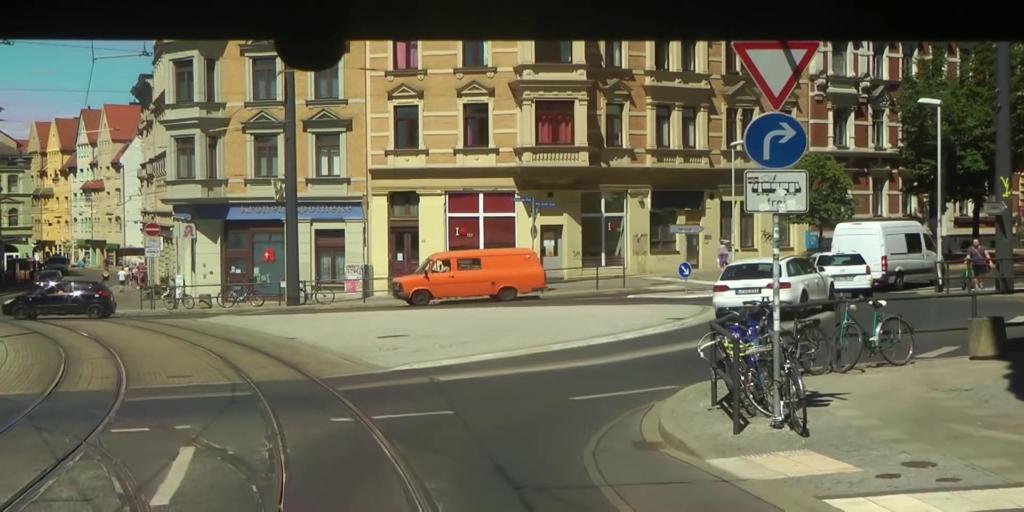}
            \caption*{Original Image}
            
        \end{subfigure}
    \end{minipage}
    \hfill
    \begin{minipage}{0.5\textwidth}
        \begin{subfigure}[b]{0.49\textwidth}
            \centering
\includegraphics[width=\textwidth]{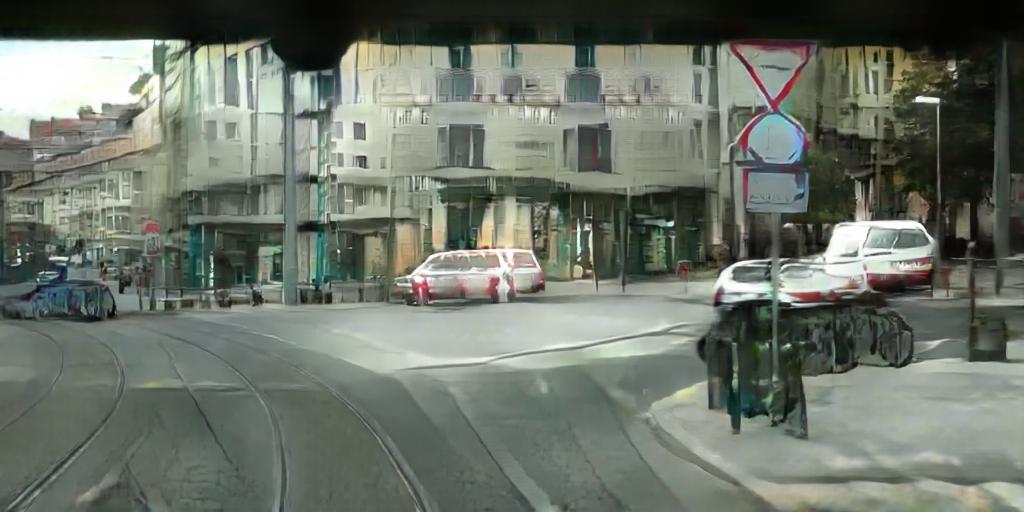}
            \caption*{Cloudy}
            
        \end{subfigure}
        \hfill
        \begin{subfigure}[b]{0.49\textwidth}
            \centering
            \includegraphics[width=\textwidth]{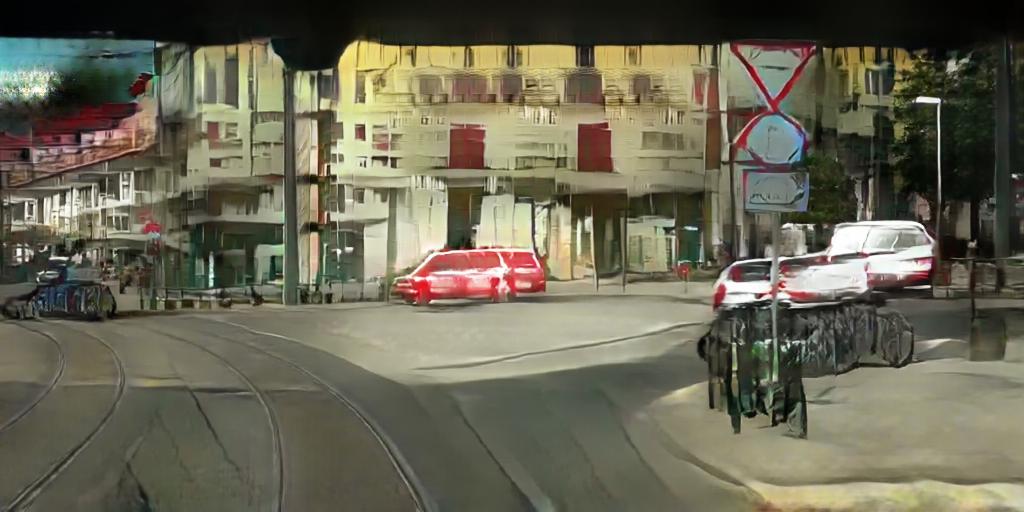}
            \caption*{Sunshine}
           
        \end{subfigure}
        \hfill
        \begin{subfigure}[b]{0.49\textwidth}
            \centering
\includegraphics[width=\textwidth]{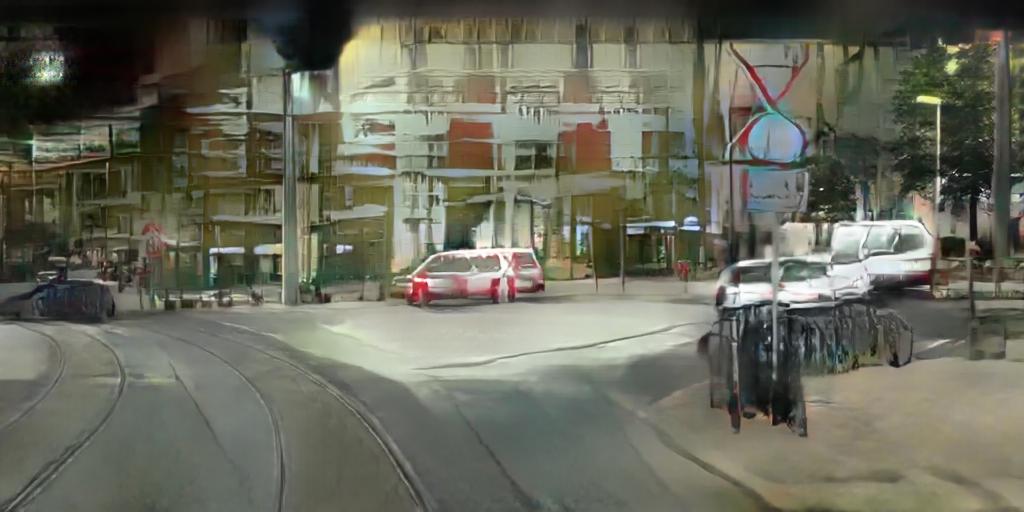}
            \caption*{Night}
            
        \end{subfigure}
        \hfill
        \begin{subfigure}[b]{0.49\textwidth}
            \centering
\includegraphics[width=\textwidth]{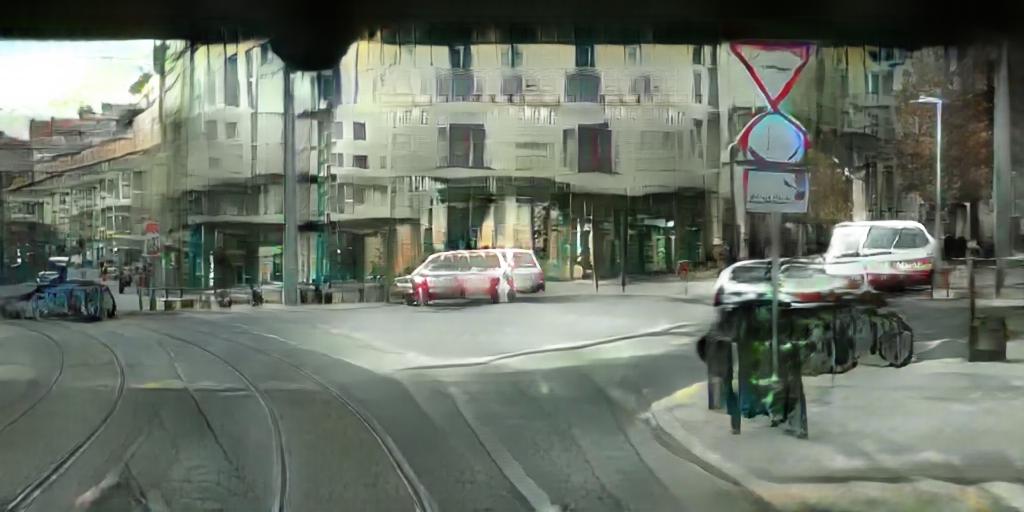}
            \caption*{Snow}
            
        \end{subfigure}
    \end{minipage}
    \caption{\textbf{Top}: Original image, and synthesized versions with minor artifacts.\\ \textbf{Bottom}: Original image, and synthesized versions with significant artifacts. }
    \label{smooth}
\end{figure}
       
\section{Results and Experiences}\label{results}
\subsubsection{Scenario Simulation}
To simulate test scenarios we trained the proposed cGAN on RailSem19 based on the default implementation provided by the authors \cite{wang2018high}. Applying k-means clustering to all style encoding indeed enables us to locate $\textit{k=}10$ distinct regions in the style space that correspond to different lighting and weather conditions. Fig. \ref{fig-1} shows some styles represented by cluster centers for the sky class, which we refer to as prototypical cloudy, sunny, and night during our experiments. To manipulate illumination, we replace the feature vector of the sky instance in a given image with desired cluster center and synthesize a new image as outlined in Section \ref{pm}. Changing the weather to snowfall involves manipulating several instances in the railway scene individually. Therefore, we replace the original style features of all semantic classes with their respective cluster centers that best depict the snowfall weather condition. Fig. \ref{snow-encoding} shows how the features of each semantic category are altered to translate an original weather condition into snowfall. Overall, images with a significant amount of sky, vegetation, terrain or rails are of high quality, e.g. Fig.\ref{smooth}. However, we also observed significant artifacts while encoding images with buildings, people and cars. Fig. \ref{smooth} shows such an example where also the simulation of snow fails. Hence, model evaluations on synthesized examples should ideally be complemented by manual human inspection on a case-by-case basis to ensure sound conclusions.
\subsubsection{Model Evaluation} 
\begin{figure}[!b]
	\centering
	\includegraphics[width=1.\textwidth]{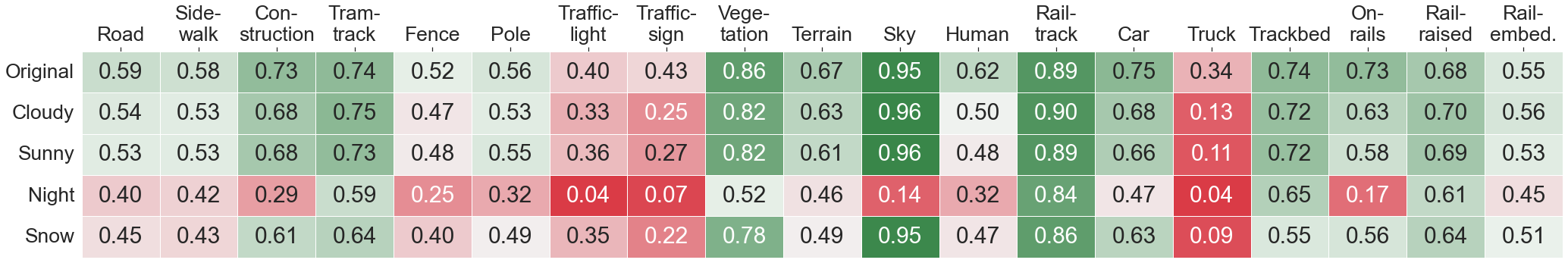}
	\caption{Class-wise IoU results of the trained segmentation model on test data}\label{classIoU}
\end{figure}

\begin{figure}[!b]
\centering
    \begin{subfigure}[t]{0.24\textwidth}
        \centering
        \includegraphics[width=\textwidth]{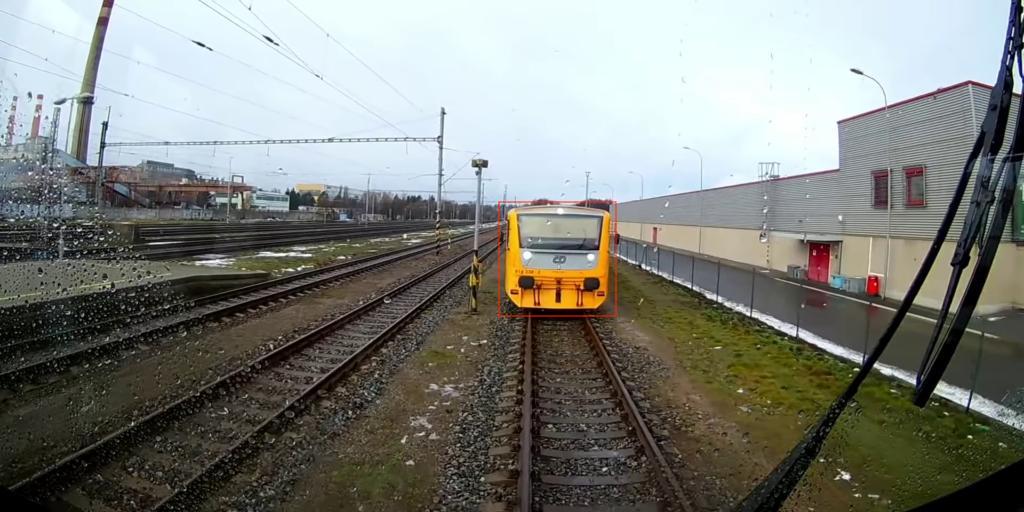}
        
    \end{subfigure}
    \hfill
    \begin{subfigure}[t]{0.24\textwidth}
        \centering
        \includegraphics[width=\textwidth]{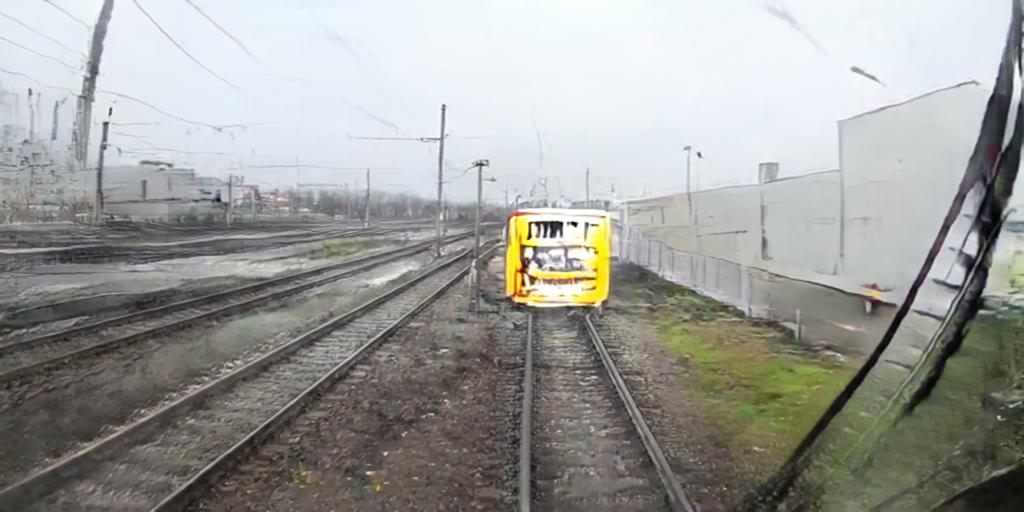}
       
    \end{subfigure}
    \hfill
    \begin{subfigure}[t]{0.24\textwidth}
        \centering
        \includegraphics[width=\textwidth]{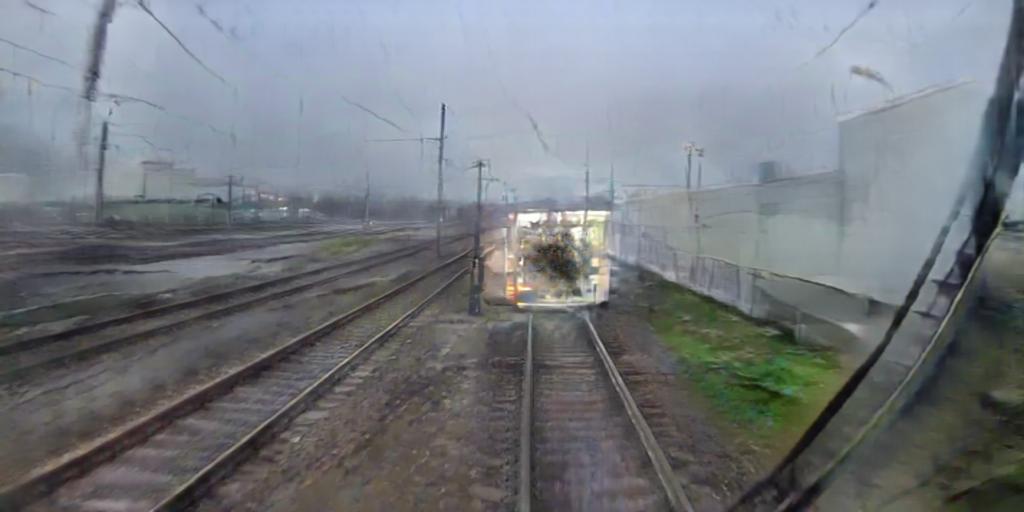}
        
    \end{subfigure}
    \hfill
    \begin{subfigure}[t]{0.24\textwidth}
        \centering
        \includegraphics[width=\textwidth]{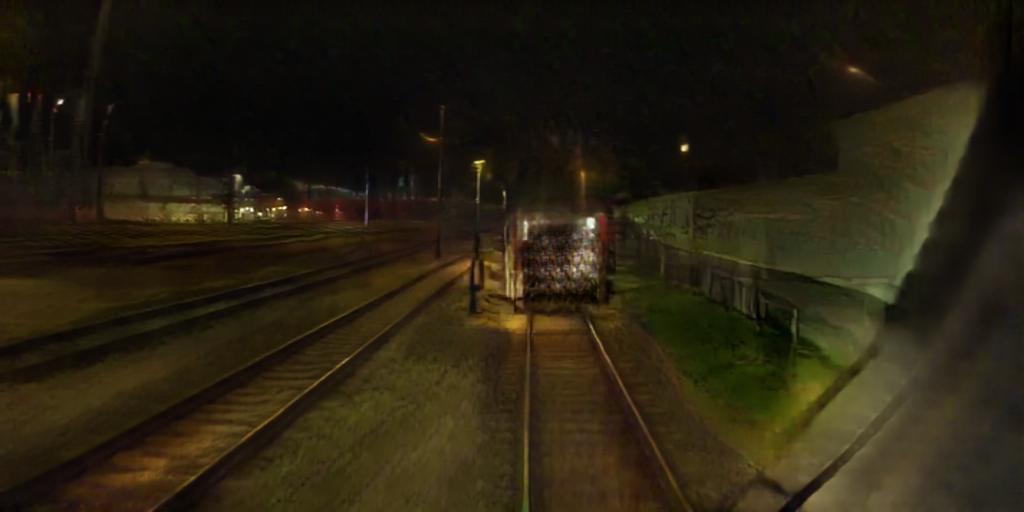}
        
    \end{subfigure}

    \begin{subfigure}[t]{0.24\textwidth}
        \centering
        \includegraphics[width=\textwidth]{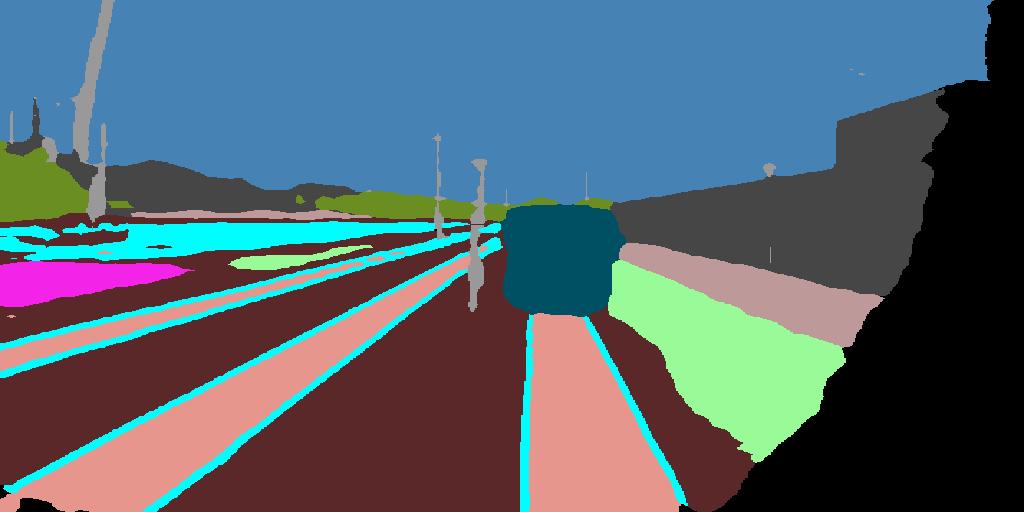}
        \caption{IoU: 0.89}
        \label{fig:y equals x}
    \end{subfigure}
    \hfill
    \begin{subfigure}[t]{0.24\textwidth}
        \centering
        \includegraphics[width=\textwidth]{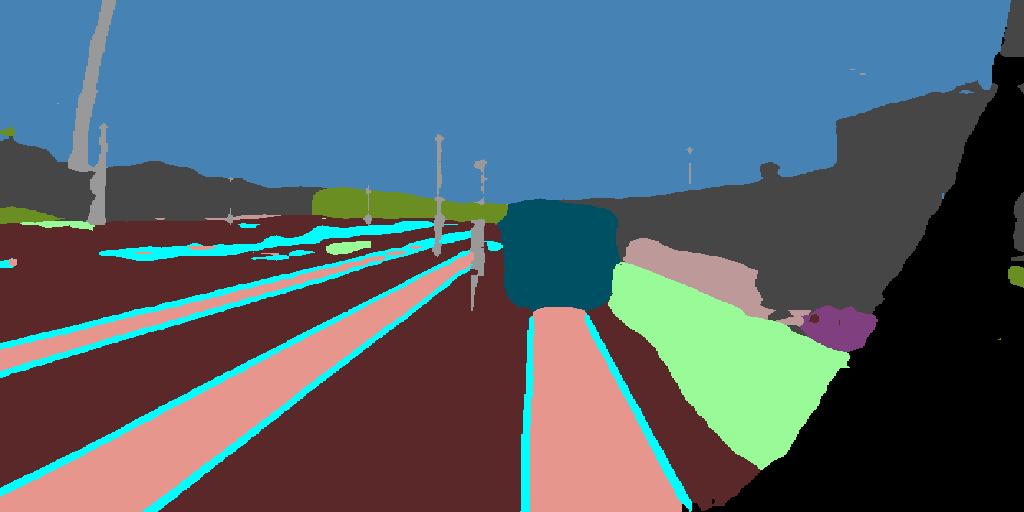}
        \caption{IoU: 0.90}
        \label{on-rail1}
    \end{subfigure}
    \hfill
    \begin{subfigure}[t]{0.24\textwidth}
        \centering
        \includegraphics[width=\textwidth]{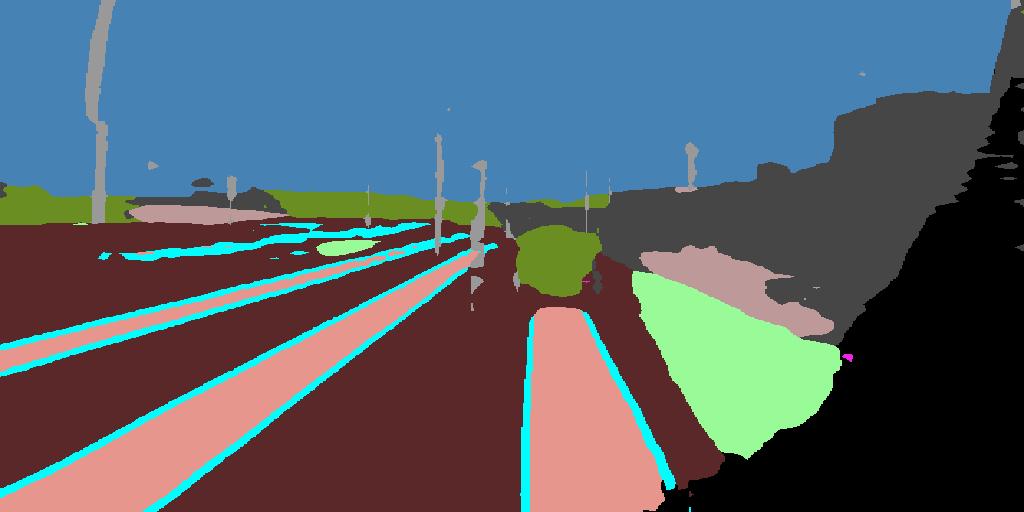}
        \caption{IoU: 0.0}
        \label{on-rail2}
    \end{subfigure}
    \hfill
    \begin{subfigure}[t]{0.24\textwidth}
        \centering
        \includegraphics[width=\textwidth]{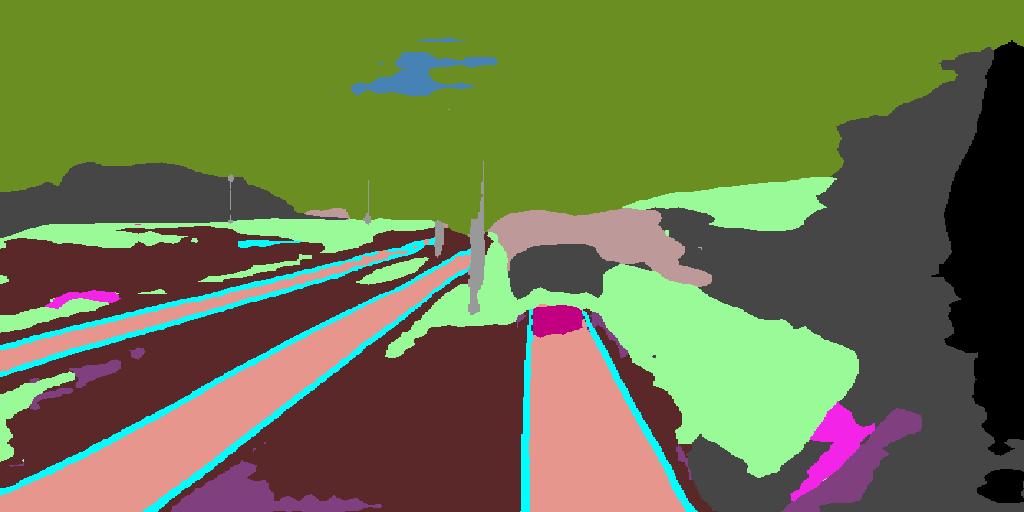}
        \caption{IoU: 0.0}
        \label{on-rail3}
    \end{subfigure}
     \caption{Change in IoU for an on-rail object when changing the original lighting to night-time. Huge performance decline when going from \ref{on-rail1} to \ref{on-rail2}.}\label{on_rail}
\end{figure}

\begin{figure}[!b]
\centering
    \begin{subfigure}[b]{0.24\textwidth}
        \centering
        \includegraphics[width=\textwidth]{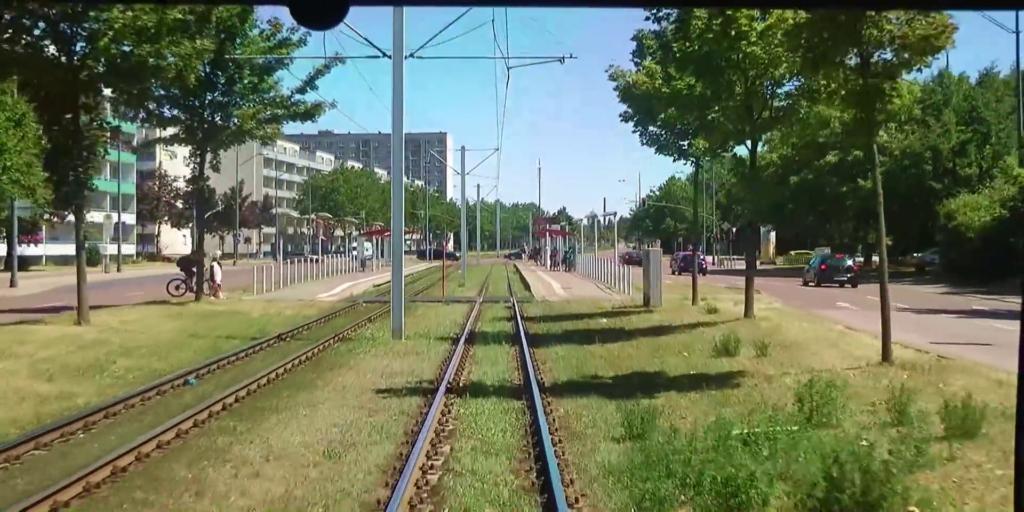}
        
    \end{subfigure}
    \hfill
     \begin{subfigure}[b]{0.24\textwidth}
        \centering
        \includegraphics[width=\textwidth]{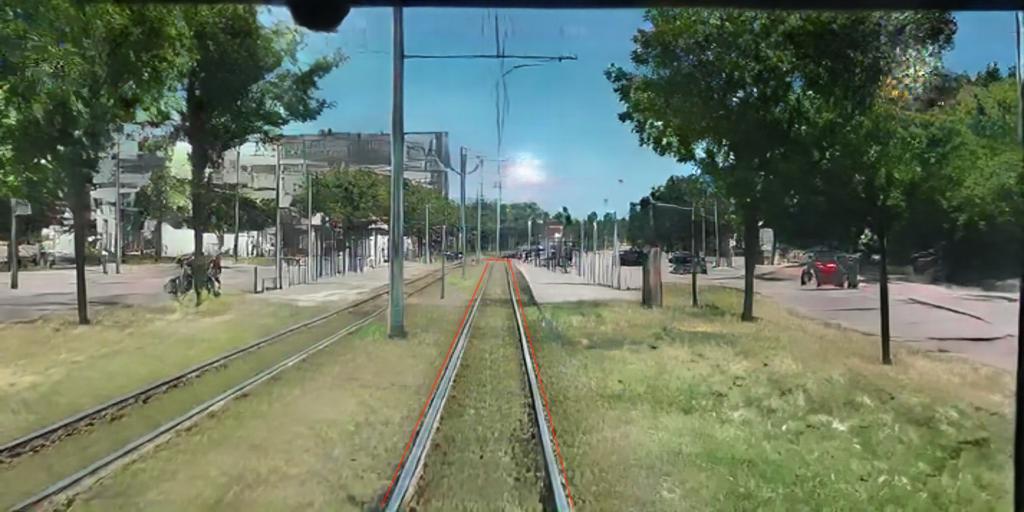}
        
    \end{subfigure}
    \hfill
    \begin{subfigure}[b]{0.24\textwidth}
        \centering
        \includegraphics[width=\textwidth]{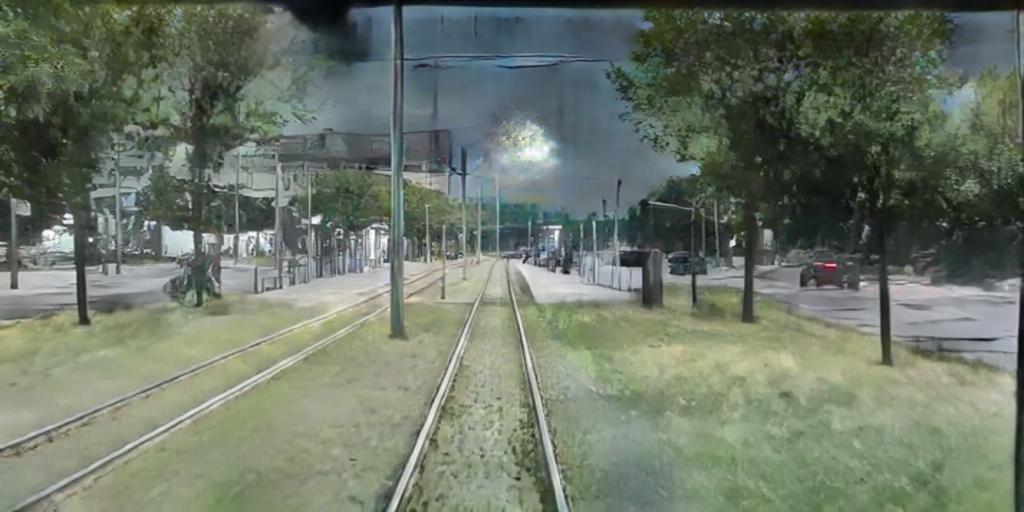}
        
    \end{subfigure}
    \hfill
    \begin{subfigure}[b]{0.24\textwidth}
        \centering
        \includegraphics[width=\textwidth]{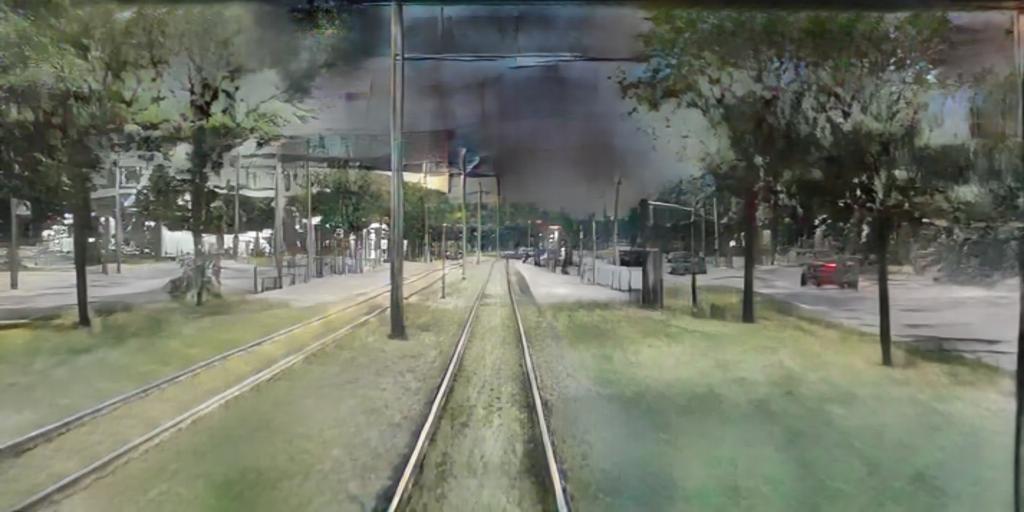}
        
    \end{subfigure}
    
    \begin{subfigure}[b]{0.24\textwidth}
        \centering
        \includegraphics[width=\textwidth]{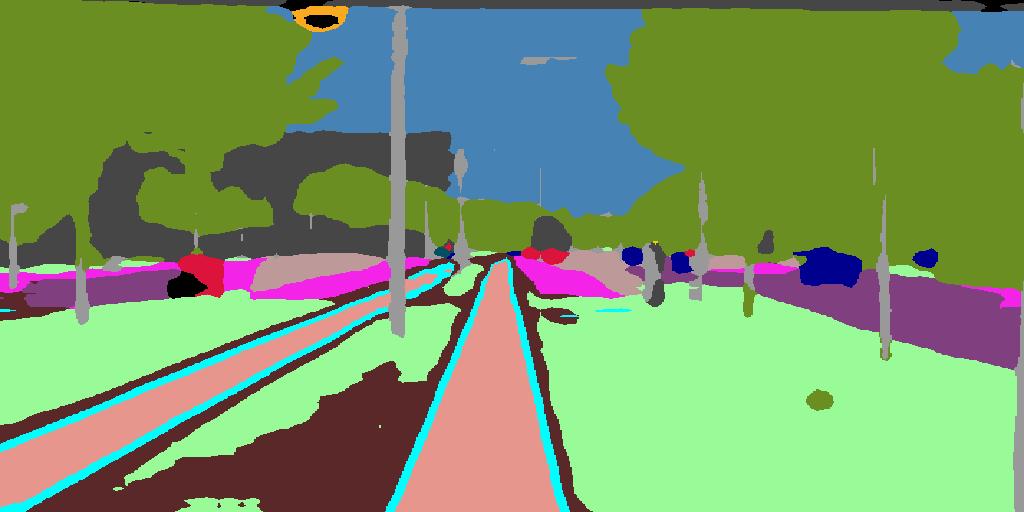}
        \caption{IoU: 0.95}
        \label{rail-track-1}
    \end{subfigure}
    \hfill
    \begin{subfigure}[b]{0.24\textwidth}
        \centering
        \includegraphics[width=\textwidth]{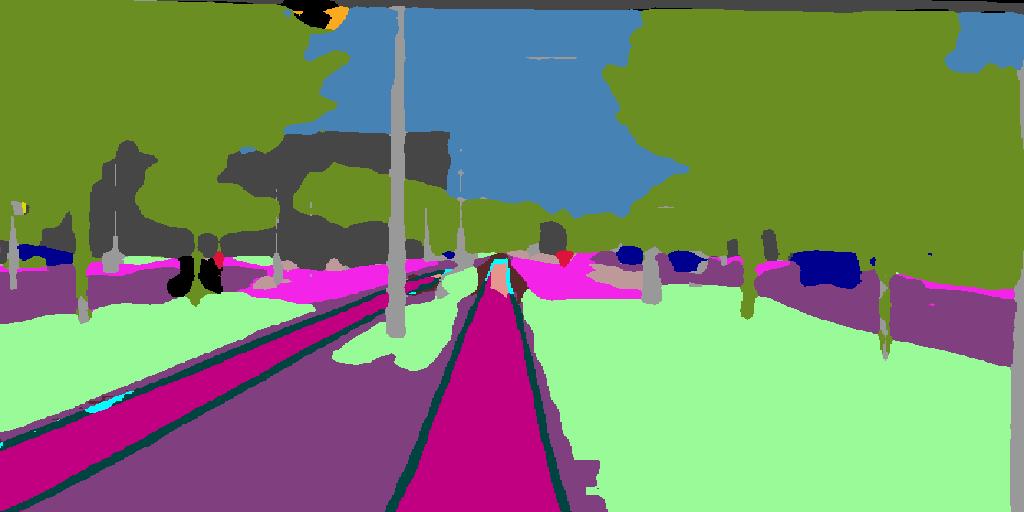}
        \caption{IoU: 0.01}
        \label{rail-track-2}
    \end{subfigure}
    \hfill
    \begin{subfigure}[b]{0.24\textwidth}
        \centering
        \includegraphics[width=\textwidth]{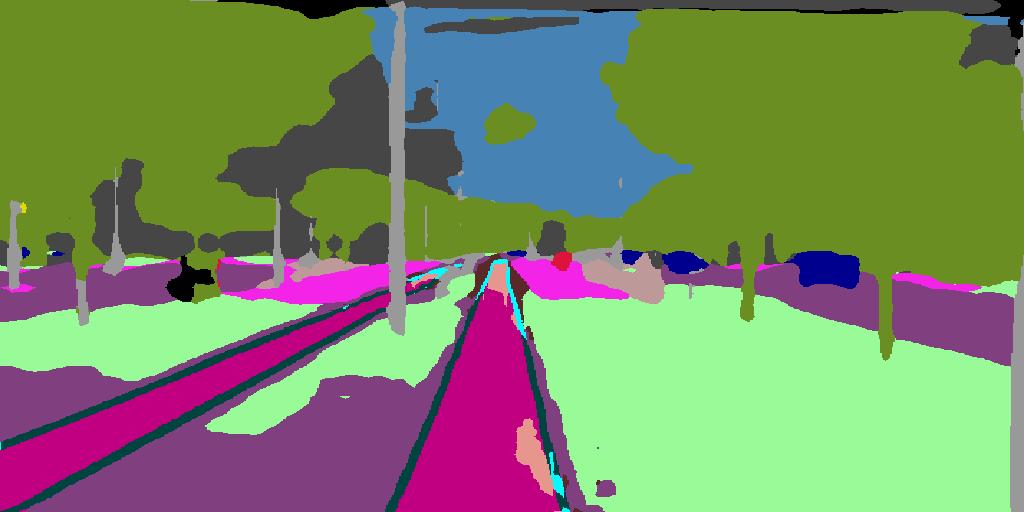}
        \caption{IoU: 0.06}
        \label{rail-track-3}
    \end{subfigure}
    \hfill
    \begin{subfigure}[b]{0.24\textwidth}
        \centering
        \includegraphics[width=\textwidth]{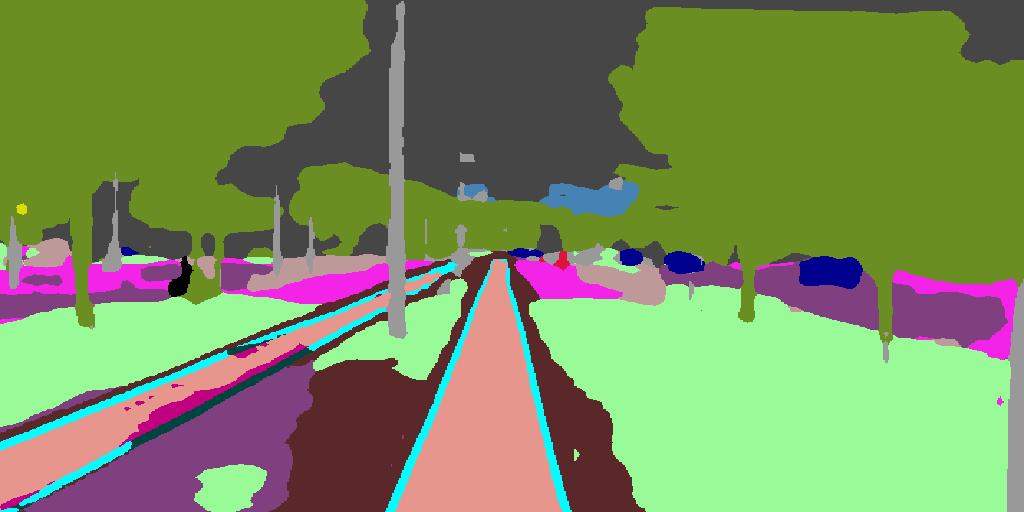}
        \caption{IoU: 0.91}
        \label{rail-track-4}
    \end{subfigure}
    \caption{Change in IoU of rail tracks when moving from original to night illumination. Unstable performance when going from \ref{rail-track-1} to \ref{rail-track-2} and \ref{rail-track-3} to \ref{rail-track-4}.}\label{sunny1}
\end{figure}

\begin{figure}[!b]
\centering
    \begin{subfigure}[t]{0.24\textwidth}
        \centering
        \includegraphics[width=\textwidth]{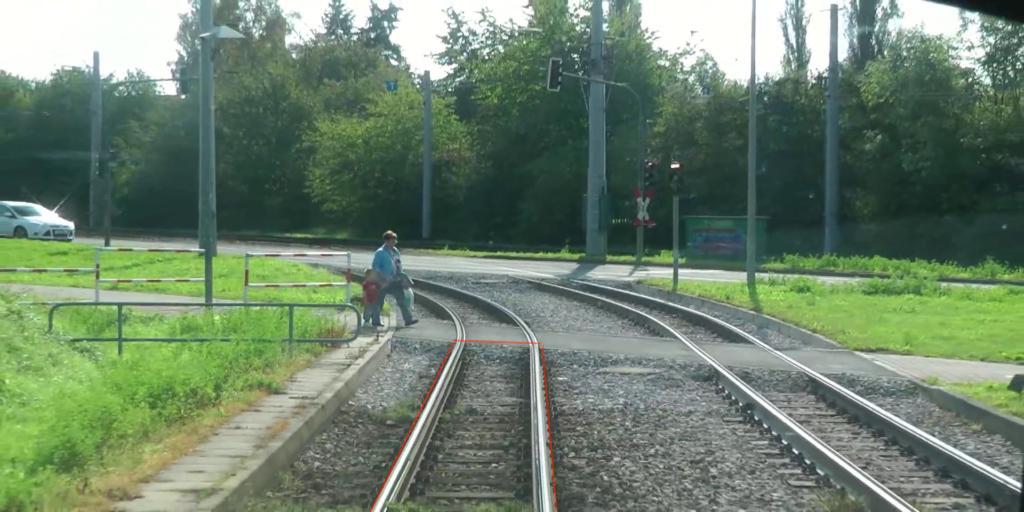}
        
    \end{subfigure}
    \hfill
    \begin{subfigure}[t]{0.24\textwidth}
        \centering
        \includegraphics[width=\textwidth]{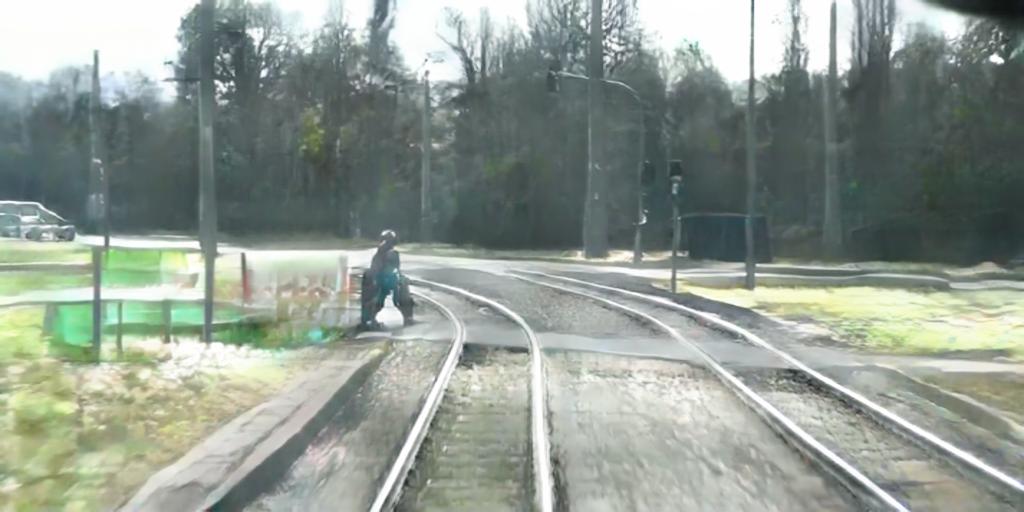}
        
    \end{subfigure}
    \hfill
    \begin{subfigure}[t]{0.24\textwidth}
        \centering
        \includegraphics[width=\textwidth]{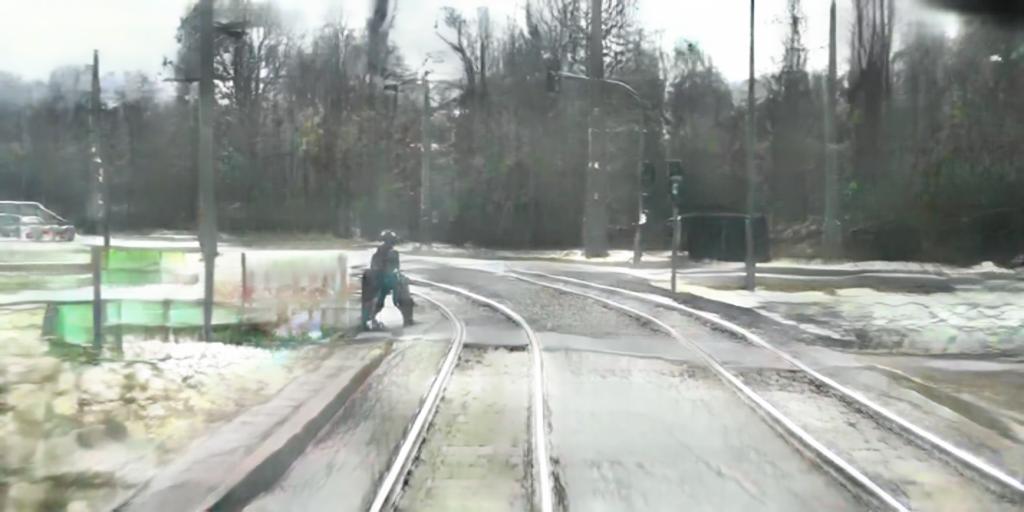}
        
    \end{subfigure}
    \hfill
    \begin{subfigure}[t]{0.24\textwidth}
        \centering
        \includegraphics[width=\textwidth]{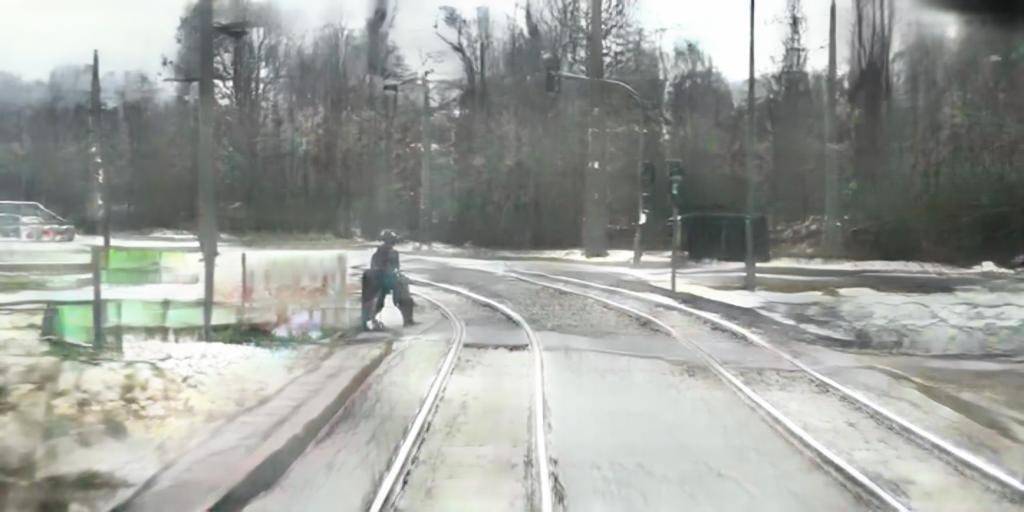}
        
    \end{subfigure}

    \begin{subfigure}[t]{0.24\textwidth}
        \centering
        \includegraphics[width=\textwidth]{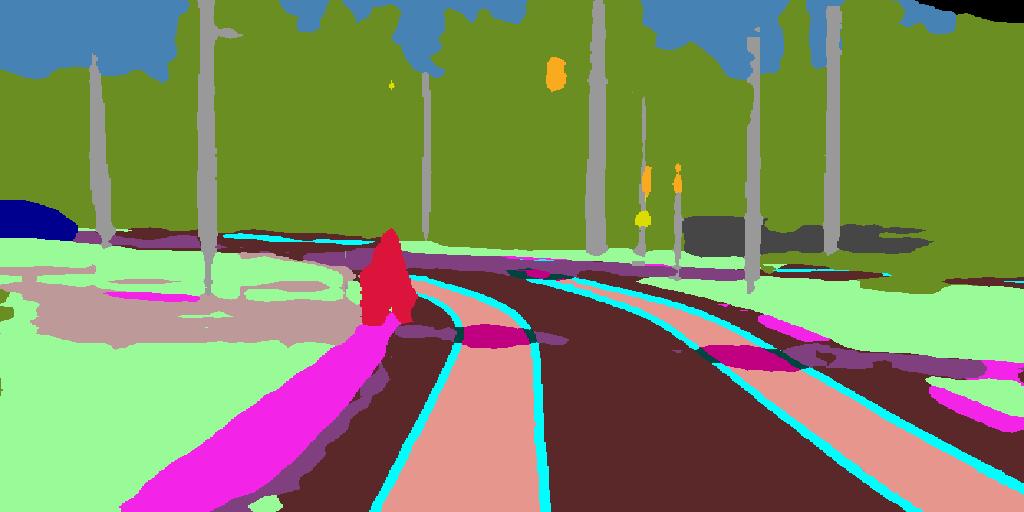}
        \caption{IoU: 0.93}
        \label{rail-track-5}
    \end{subfigure}
    \hfill
    \begin{subfigure}[t]{0.24\textwidth}
        \centering
        \includegraphics[width=\textwidth]{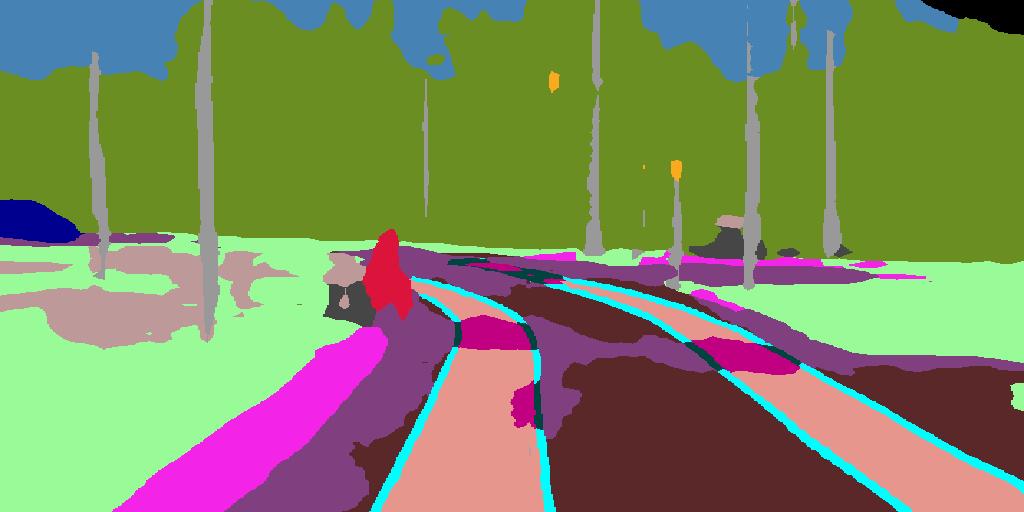}
        \caption{IoU: 0.93}
        \label{rail-track-6}
    \end{subfigure}
    \hfill
    \begin{subfigure}[t]{0.24\textwidth}
        \centering
        \includegraphics[width=\textwidth]{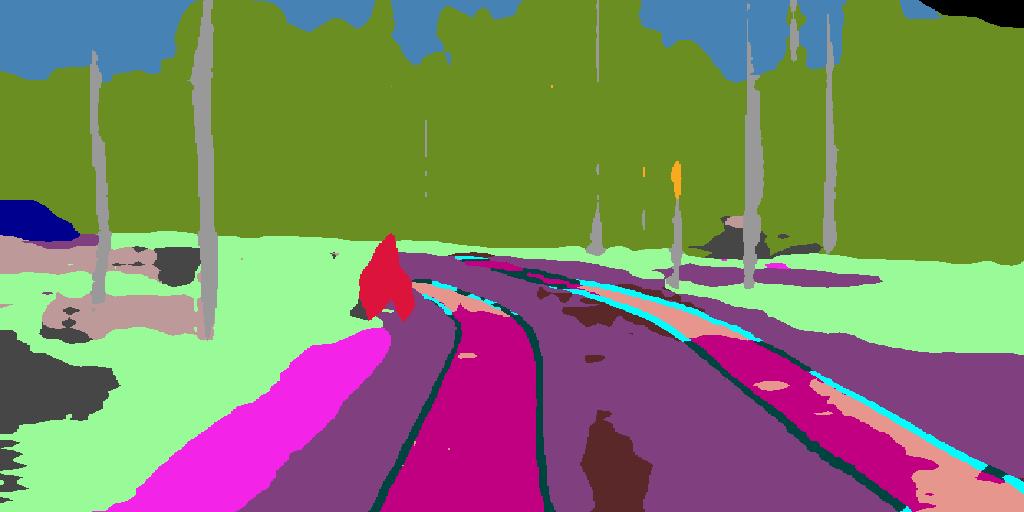}
        \caption{IoU: 0.26}
        \label{rail-track-7}
    \end{subfigure}
    \hfill
    \begin{subfigure}[t]{0.24\textwidth}
        \centering
        \includegraphics[width=\textwidth]{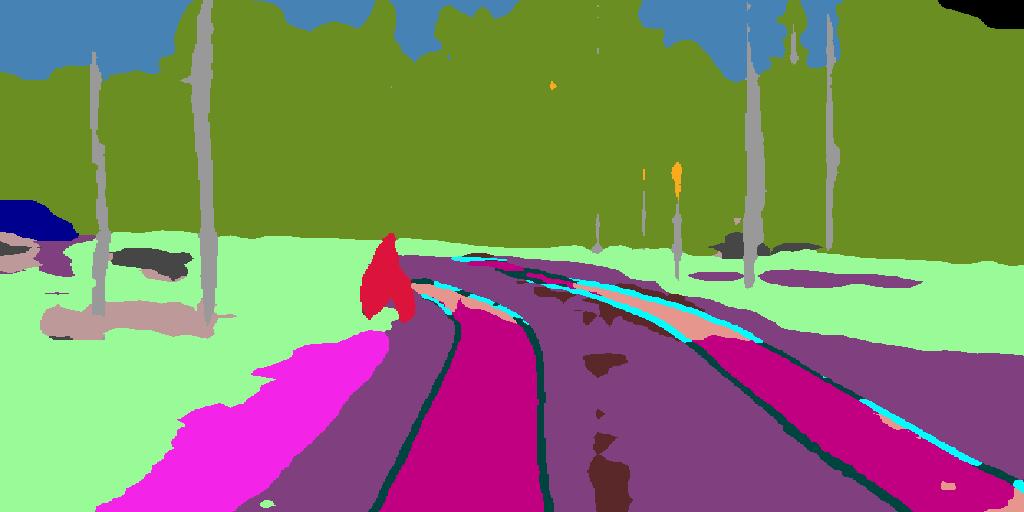}
        \caption{IoU: 0.10}
        \label{rail-track-8}
    \end{subfigure}
    \caption{Change in IoU of the rail tracks when changing the original weather condition to snow. Despite similarity, performance drops from \ref{rail-track-6} to \ref{rail-track-7}.}\label{snow1}
\end{figure}

For our experiments we use the PSPNet \cite{zhao2017pyramid}, which we train on RailSem19 similarly to the procedure in \cite{zendel2019railsem19}. Out of the 8500 available images we randomly selected 7140 for training and fine-tuning leaving us only 1360 for rigorous testing. On this test set, we achieve a mean IoU of 0.65 over all classes which is comparable with the reference performance reported in \cite{zendel2019railsem19}. To validate if the model also complies with the ODD-related requirements of proper operation under different lighting conditions and snow we applied our proposed methodology to create 4 new versions of the original test set where we modified the style of all images accordingly. The corresponding IoU scores per segmentation class are reported in Fig. \ref{classIoU}. Our evaluation reveals that in all scenarios the model performs well with respect to the detection of tram/rail tracks or trackbeds but seems to struggle with traffic lights/signs or trucks. Also, simulating nighttime conditions seems to be particularly detrimental to the model performance, as for instance indicated by the significant IoU drops for segmenting cars, humans, construction sites or other on-rail vehicles. Since accurate detection of corresponding objects is potentially safety-critical our evaluation possibly reveals a crucial deficiency. To verify if this is indeed the case or just due to simulation artifacts we can also evaluate the model behavior on individual examples transitioning from their original style to night mode. Fig. \ref{on_rail} displays an image with an on-rail vehicle in front of the train, that is accurately detected under the original illumination. Progressively moving to night causes the model to miss the object, but its visual appearance also becomes unnatural requiring closer inspection by a human auditor. Moreover, by evaluating other images under style transition we can demonstrate other deficiencies. In Fig. \ref{sunny1} the model performs well on the original image. Since it is already sunny, the synthesized sunny version is quite similar but the rail tracks are perceived as tram tracks by the model. Surprisingly, when transitioning towards night conditions the prediction suddenly turns correct at some point, although the visual appearance of the rails changes only marginally. Similarly in Fig. \ref{snow1}, moving to snow causes the model to suddenly confuse rail and tram tracks despite the high visual similarity of the tracks in all pictures.

\section{Conclusion} \label{conc}
In this work we report our experiences with cGANs to validate if an AI-powered model complies with typical ODD requirements, especially varying weather and lighting conditions. We intend to expand the approach to also enable the rendering of new objects such as obstacles, persons or vehicles on the rails. Comparing the simulation quality of similar generative model types, such as variants of recently popularized Diffusion Models \cite{kawar2022imagic} is also relevant for future work.


\subsubsection*{Acknowledgement} We acknowledge the support from the Federal Ministry for Economic Affairs and Climate Action (BMWK) via grant agreement 19I21039A.
 
\bibliographystyle{splncs04}
\bibliography{ref.bib}

\end{document}